\documentclass[11pt]{article}
\usepackage{aaai2026}
\usepackage{times}  
\usepackage{helvet}  
\usepackage{courier}  
\usepackage[hyphens]{url}  
\usepackage{graphicx} 
\usepackage[numbers,sort&compress]{natbib}
\usepackage{caption} 
\usepackage[ruled,vlined]{algorithm2e}
\usepackage{hyperref}
\hypersetup{
  colorlinks=true,
  urlcolor=blue,   
}
\usepackage{fontawesome5} 
\usepackage{newfloat}
\usepackage{listings}
\DeclareCaptionStyle{ruled}{labelfont=normalfont,labelsep=colon,strut=off} 
\usepackage[T1]{fontenc}
\usepackage{newtxtext,newtxmath} 
\nocopyright

\title{ASCD: Attention-Steerable Contrastive Decoding for Reducing Hallucination in MLLM}
\author{
    Yujun Wang\textsuperscript{\rm 1},
    Aniri\textsuperscript{\rm 2},
    Jinhe Bi\textsuperscript{\rm 2},
    Soeren Pirk\textsuperscript{\rm 1},
    Yunpu Ma\textsuperscript{\rm 2,3}
}
\affiliations{
    \textsuperscript{\rm 1}Christian-Albrechts-Universität zu Kiel (CAU)\\
    \textsuperscript{\rm 2}Ludwig Maximilian University of Munich\\
    \textsuperscript{\rm 3}Munich Center for Machine Learning\\
    \faGithub\; \href{https://github.com/BroJunn/ASCD}{https://github.com/BroJunn/ASCD}

}

\usepackage{bibentry}

\usepackage{amsmath}
\usepackage{algpseudocode}
\usepackage{multirow} 
\usepackage{xcolor}
\usepackage{colortbl}
\usepackage{booktabs}
\newcommand{\eg}{\emph{e.g.}}
\usepackage{makecell}
\usepackage{placeins}

\usepackage{amsfonts}

\begin{document}

\maketitle

%



\begin{abstract}
Multimodal large language models (MLLMs) frequently hallucinate by over-committing to spurious visual cues. Prior remedies—Visual and Instruction Contrastive Decoding (VCD, ICD)—mitigate this issue, yet the mechanism remains opaque. We first empirically show that their improvements systematically coincide with \emph{redistributions of cross-modal attention}. Building on this insight, we propose \textbf{Attention-Steerable Contrastive Decoding (ASCD)}, which \emph{directly steers the attention scores during decoding}. ASCD combines (i) \textit{positive steering}, which amplifies automatically mined \textit{text-centric heads}—stable within a model and robust across domains—with (ii) \textit{negative steering}, which dampens on-the-fly identified critical visual tokens. The method incurs negligible runtime/memory overhead and requires no additional training. Across five MLLM backbones and three decoding schemes, ASCD reduces hallucination on \textsc{POPE}, \textsc{CHAIR}, and \textsc{MMHal-Bench} by up to 38.2\% while \emph{improving} accuracy on standard VQA benchmarks, including \textsc{MMMU}, \textsc{MM-VET}, \textsc{ScienceQA}, \textsc{TextVQA}, and \textsc{GQA}. These results position attention steering as a simple, model-agnostic, and principled route to safer, more faithful multimodal generation.
\end{abstract}

\section{Introduction}
\label{sec:intro}

Recent advances in large language models (LLMs) 
 \citep{yang2024qwen2technicalreport, touvron2023llamaopenefficientfoundation, abdin2024phi3technicalreporthighly, raffel2023exploringlimitstransferlearning, brown2020languagemodelsfewshotlearners, devlin2019bertpretrainingdeepbidirectional} have led to impressive results in a wide array of natural language processing tasks. Building on these successes, researchers have extended LLMs with visual inputs that enable multimodal large language models (MLLMs) such as LLaVA \citep{liu2023llava, liu2024llavanext}. These MLLMs can handle complex tasks like image captioning \cite{anderson2018bottomuptopdownattentionimage}, visual question answering \cite{agrawal2016vqavisualquestionanswering}, and multimodal dialogue \cite{das2017visualdialog}. Existing approaches \citep{dai2023instructblipgeneralpurposevisionlanguagemodels, liu2023llava, liu2024llavanext, zhou2024tinyllavaframeworksmallscalelarge, chen2023palijointlyscaledmultilinguallanguageimage, alayrac2022flamingovisuallanguagemodel, bi2024visual} show remarkable potential to bridge the gap between vision and language.

Despite these achievements, MLLMs often inherit a critical limitation from LLMs: the tendency to produce \emph{hallucinations} \citep{huang2024visualhallucinationsmultimodallarge, bai2024hallucinationmultimodallargelanguage, liu2024surveyhallucinationlargevisionlanguage}. These hallucinations arise when a model over-relies on partial or misleading cues, generating responses that are incorrect or do not correspond to the provided input. 


\begin{figure}[t]
  \centering
   \includegraphics[width=1.0\linewidth]{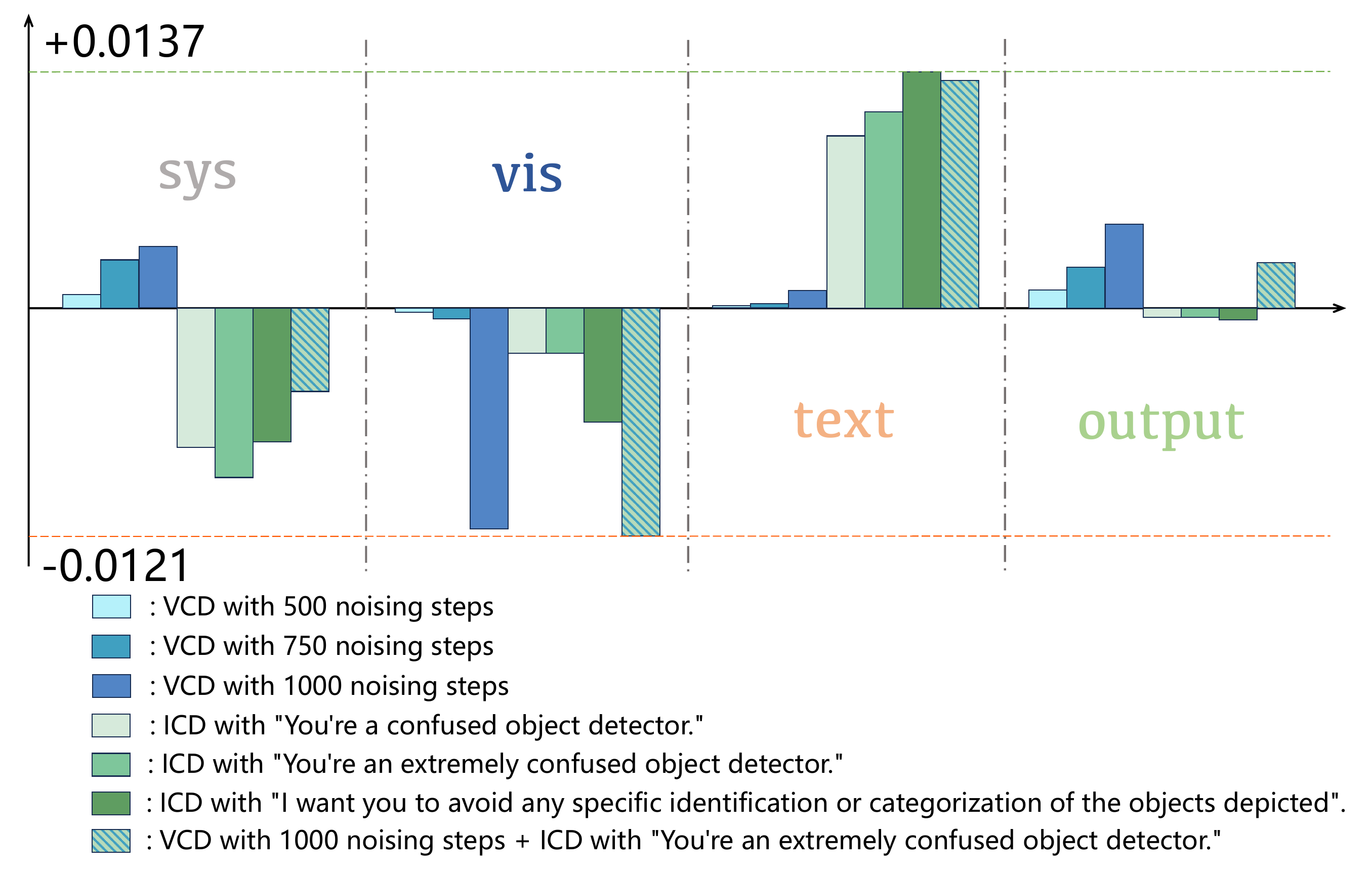}
   \caption{Impact of VCD and ICD on attention distribution.
On 500 COCO images, we measure how Visual~(VCD) and Instruction~(ICD) Contrastive Decoding redistribute
attention in LLaVA-1.5.  Both techniques—and their combination—lower attention on visual tokens (\textit{vis}) while raising it on textual tokens (\textit{text}), with stronger perturbations yielding larger shifts. This suggests that the reduction in hallucinations achieved by VCD and ICD is largely attributable to the attendant shifts in cross-modal attention, rather than to the logit-subtraction step alone.}
   \label{fig:motiv-1}
\end{figure}

To mitigate hallucinations, two general strategies have emerged: \emph{training-phase} interventions and \emph{inference-phase} interventions. 
In the training phase, auxiliary supervision \cite{chen2023mitigatinghallucinationvisuallanguage} or reinforcement learning \cite{benkish2024mitigatingopenvocabularycaptionhallucinations} can help align model outputs with factual or human-preferred references. However, these approaches require additional data or complex reward modeling, which may be costly or infeasible in certain scenarios. In contrast, \emph{inference-phase} methods \citep{zhou2024tinyllavaframeworksmallscalelarge, zhao2024mitigatingobjecthallucinationlarge, deng2024seeingbelievingmitigatinghallucination, wang2024mitigatinghallucinationslargevisionlanguage, leng2023mitigatingobjecthallucinationslarge} aim to correct or filter erroneous outputs without retraining. \emph{Contrastive decoding} is particularly appealing as it leverages negatively perturbed or prefixed inputs to steer the model away from hallucinations in a training-free manner. Two notable recent methods for contrastive decoding are Visual Contrastive Decoding (VCD) \cite{leng2023mitigatingobjecthallucinationslarge} that perturbs an input image (\eg, via noising) to generate a ``negative result'' of logits, which is then subtracted from the original logits to suppress hallucinations, and Instruction Contrastive Decoding (ICD) \cite{wang2024mitigatinghallucinationslargevisionlanguage} that prepends a negative prefix to the prompt (\eg, ``You are a confused object detector'') to generate a signal that shifts the model’s predictions away from hallucinated content. Both methods offer a lightweight, yet effective approach to reducing hallucinations. However, upon closer examination, we find that these methods construct contrasting branches through surface-level modifications—either perturbing the image (VCD) or prefixing the prompt (ICD)—without explicitly addressing the underlying cause of hallucinations. 
\emph{Attention steering} like OPERA, IBD and PAI \citep{liu2024payingattentionimagetrainingfree, zhu2024ibdalleviatinghallucinationslarge, huang2024operaalleviatinghallucinationmultimodal} is also a common inference-phase remedy to reduce hallucinations. However, PAI introduces the notion of ``text inertia''—the tendency of an MLLM to keep generating text-driven content even when the image is removed—but does not articulate why steering the attention matrix is the necessary lever to overcome this inertia. 

To motivate our approach, we first quantify how VCD and ICD reshape a model’s internal attention. As evidenced by Fig.~\ref{fig:motiv-1}, both techniques produce a systematic reallocation of attention from visual tokens to textual tokens. This insight raises a natural question: \emph{why not directly steer the attention mechanism itself?} To this end, we propose an \emph{Attention-Steerable Contrastive Decoding (ASCD)} framework to manipulate attention. Specifically, the attention modification is integrated into a contrastive decoding pipeline to both enhance visual cues and suppress negative signals. We further develop a dynamic head-selection mechanism to identify ``text-centric'' heads that disproportionately focus on textual cues, enabling more targeted positive adjustments. In parallel, we introduce a complementary mechanism that restricts negative steering to only the most critical visual tokens, ensuring that suppression is applied solely where necessary to mitigate hallucinations while preserving essential visual details. In summary, our contributions are as follows:
(1)~We analyze how recent contrastive decoding methods (VCD, ICD) create ``negative samples'' that fundamentally alter attention;
(2)~We propose an \emph{attention-steerable contrastive decoding} method that explicitly modulates attention distributions to offer a more principled way to mitigate hallucinations in the inference phase;
(3)~We faithfully reproduce VCD and ICD to ensure fair comparison with prior work. Across five representative MLLM backbones (LLaVA-1.5 7B, LLaVA-NeXT 7B, Phi2-SigLIP, LLaVA-1.5 13B and Qwen2.5-VL-Instruct), three decoding schemes (greedy, nucleus, and beam search), and three hallucination-focused benchmarks \citep{rohrbach2019objecthallucinationimagecaptioning, li2023evaluatingobjecthallucinationlarge, sun2023aligninglargemultimodalmodels} (\textsc{POPE}, \textsc{CHAIR}, \textsc{MMHal-Bench}), our approach consistently reduces hallucinations and strengthens visual grounding. At the same time, it improves performance on standard VQA benchmarks \citep{yue2024mmmumassivemultidisciplinemultimodal, yu2024mmvetevaluatinglargemultimodal, lu2022learnexplainmultimodalreasoning, singh2019vqamodelsread, hudson2019gqanewdatasetrealworld}, including \textsc{MMMU}, \textsc{MM-VET}, \textsc{ScienceQA}, \textsc{TextVQA}, and \textsc{GQA} whereas other methods suffer from degraded performance on these benchmarks.



\begin{figure*}
  \centering
  \includegraphics[width=1.0\linewidth]{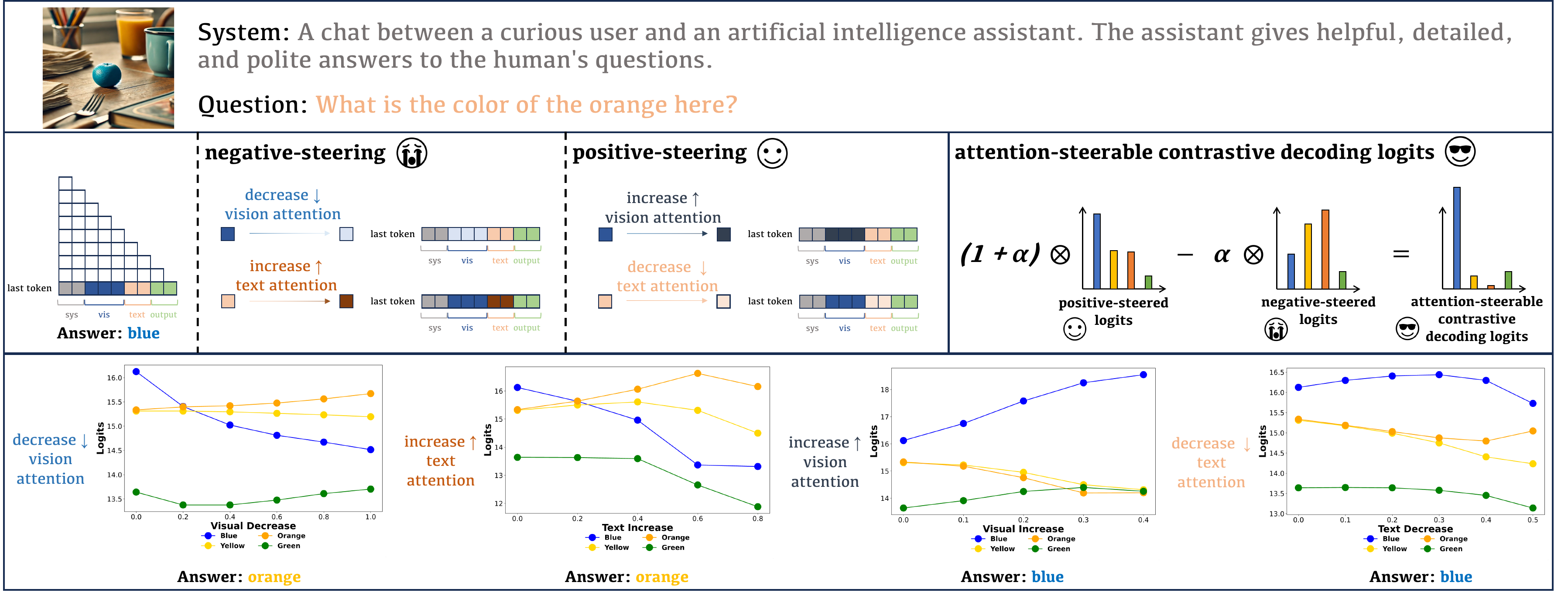}
  \caption{A motivating example of proactive attention steering in a visually ambiguous scenario. Top: Conversation context in which the “orange” appears blue-tinted. Middle: Effects of \emph{negative steering} (decrease vision attention / increase text attention) and \emph{positive steering} (increase vision attention / decrease text attention); ASCD contrasts the two steered logits to suppress hallucination and produce the perception-consistent answer. Bottom: Color-token logits change with the steering strength for visual and textual attention, corresponding to the steering above.
}
  \label{fig:motiv-2}
\end{figure*}
\section{Related Work}
\label{sec:related_work}

\textbf{Multimodal Large Language Models.}  
Multimodal Large Language Models (MLLMs) have significantly advanced the field of artificial intelligence by integrating vision and language understanding, enabling a wide range of vision-language tasks \citep{dai2023instructblipgeneralpurposevisionlanguagemodels, zhu2023minigpt4enhancingvisionlanguageunderstanding, liu2024llavanext, liu2023llava, bai2025qwen25vltechnicalreport, zhou2024tinyllavaframeworksmallscalelarge,rong2025backdoor,Chen_2025_CVPR}. These models typically follow a two-stage training paradigm: (1) large-scale pretraining on web-scale image-text pairs \citep{liu2023llava, li2023blip2bootstrappinglanguageimagepretraining} to learn cross-modal representations, and (2) visual instruction tuning \citep{liu2023improvedllava,bi2025prism} on task-specific datasets to enhance multimodal instruction-following capabilities. While this paradigm has led to substantial improvements in vision-language reasoning, MLLMs still face key challenges, such as hallucination -- where the model generates content that is inconsistent with the given visual input. \citep{huang2024visualhallucinationsmultimodallarge, bai2024hallucinationmultimodallargelanguage, liu2024surveyhallucinationlargevisionlanguage}.

\textbf{Mitigating Hallucinations in MLLMs.} 
Hallucinations in MLLMs are particularly pronounced in open-ended generation tasks, where models may produce content that is not aligned with the provided visual input \citep{huang2024operaalleviatinghallucinationmultimodal, jing2024faithscorefinegrainedevaluationshallucinations, zhang2023languagemodelhallucinationssnowball}. Some approaches focus on the mitigation of data bias, scaling-up of vision resolution, and alignment optimization. \citet{lovenia2024negativeobjectpresenceevaluation} introduce a technique that mines 95,000 negative samples by replacing original categories, attributes, or quantity information with similar but incorrect alternatives. This fine-grained approach effectively enriches the contrastive signal during training, thereby enhancing the model's robustness. \citet{chen2024internvlscalingvisionfoundation} propose InternVL, which scales the vision encoder up to 6 billion parameters and processes images with widths ranging from 1,664 to 6,144 pixels. While this method improves visual detail and alignment, it requires significant computational resources for pretraining with large-scale data. \citet{sun2023aligninglargemultimodalmodels} employ Reinforcement Learning from Human Feedback (RLHF) \cite{stiennon2022learningsummarizehumanfeedback} to align different modalities during training. This optimization strategy leads to a reduction in hallucinations by better integrating visual and textual cues. \citet{bi2024visual} propose a representation steering method that effectively mitigates hallucination in multimodal models.

\textbf{Contrastive Decoding Approaches.} Recent work has explored contrastive decoding as an effective, training-free means to mitigate hallucinations \cite{xiao2025mitigatingposteriorsalienceattenuation}. For instance, \citet{leng2023mitigatingobjecthallucinationslarge} introduced Visual Contrastive Decoding (VCD), which perturbs the input image to generate a negative logit branch that is subtracted from the original predictions, while \citet{wang2024mitigatinghallucinationslargevisionlanguage} employs a negative prompt to steer outputs away from hallucinated content. \citet{huo2024selfintrospectivedecodingalleviatinghallucinations} leverage a Context and Text-aware Token Selection (CT2S) strategy to selectively retain the most informative vision tokens in early decoder layers, thereby amplifying beneficial multimodal context and suppressing spurious hallucinations.
\section{Preliminaries}

Modern MLLMs integrate text and visual inputs based on powerful encoders that enable the merging of the modalities into a unified representation that is processed by a multi-layer Transformer.
While these models enable the production of coherent responses, they heavily rely on internal attention mechanisms that dictate how visual and textual cues are combined.
As discussed in previous section, subtle variations in these attention distributions can significantly impact the generated output. This observation motivates our approach: by explicitly modulating attention, we aim to enhance visual grounding and mitigate hallucinations.

\subsection{MLLM Formulation}
\label{subsec:mlm_formulation}

We consider a multimodal large language model (MLLM) that processes an image $\mathbf{I}$ and a text prompt $\mathbf{x} = \{x_1,\dots,x_N\}$ to generate an output sequence $\mathbf{y} = \{y_1,\dots,y_M\}$ in an autoregressive manner. Let $\theta$ denote the model parameters. Formally, the model maximizes:
\begin{equation}
    \mathbf{y}^* = \arg\max_{\mathbf{y}} \prod_{t=1}^{M} p_\theta\Bigl(y_t \,\Big\vert\, \mathbf{I}, \mathbf{x}, y_{<t}\Bigr),
    \label{eq:argmax}
\end{equation}
where $y_{<t}$ denotes all previously generated tokens.

\paragraph{Transformer Backbone.}
The input is processed by $L$ Transformer blocks, and
each block contains $H$ attention heads.  
We denote the unnormalized attention score matrix of the head $h$ in layer $l$ by $\mathbf{A}^{(l)}_h$.

\subsection{Proactive Steering of Attention}
\label{subsec:attn-steering-motivation}

In Figure~\ref{fig:motiv-1}, we show how visual contrastive decoding (VCD) and instruction contrastive decoding (ICD) indirectly alter attention distributions. Building on this insight, we now ask: \emph{what if we explicitly steer the model’s attention?} Figure~\ref{fig:motiv-2} provides a motivating example, illustrating how actively modulating attention can influence the final logit distribution.

Consider a simple query: \emph{``What is the color of the orange here?''} The conversation context (Figure~\ref{fig:motiv-2}) is based on LLaVA-1.5 7B, with a provided image in which the “orange” fruit appears to be tinted blue. We experiment with two distinct attention-steering scenarios: \emph{negative-steered logits} and \emph{positive-steered logits}. In each case, we proportionally adjust the visual or textual attention before finalizing the output distribution.

In the \emph{negative-steered} branch, we reduce attention to visual tokens or boost attention to the textual tokens. As shown in the histogram of logits, the model reduces its reliance on the visual input, causing it to fall back more heavily on the LLM’s inherent priors. As a result, it is more likely to generate answers that align with typical linguistic associations rather than the actual content of the image—insisting that the color is “orange”. Conversely, the \emph{positive-steered} branch increases attention to visual tokens or downgrades textual tokens, making the model more sensitive to the actual (albeit unexpected) color in the image. This leads the model to answer “blue” with higher probability.

In addition to these unidirectional adjustments, we further integrate \emph{attention steering} into the contrastive decoding framework. Instead of using the original logits for the positive branch directly (as in VCD or ICD), we inject the attention-modulated logits. Mathematically, we redefine the contrastive decoding formulation by replacing the original logits adjustment with a positively steered version:
\begin{equation}
    p_\theta^\text{final} 
    \;=\;
    (1+\alpha) p_\theta^\text{pos-steered} 
    - \alpha p_\theta^\text{neg-steered},
    \label{eq:attn_steered_cd}
\end{equation}
where $p_\theta^\text{pos-steered}$ and $p_\theta^\text{neg-steered}$ represent the output logits modified by positively or negatively steered attention.

By integrating contrastive decoding with explicit attention manipulation, our attention-steerable contrastive decoding framework (Figure~\ref{fig:motiv-2} right) sharpens the output distribution, enhances the likelihood of the correct response, while reducing the impact of competing distractors.

\section{Methodology}
\label{sec:method}

\begin{figure}[t]
  \centering
   \includegraphics[width=1.0\linewidth]{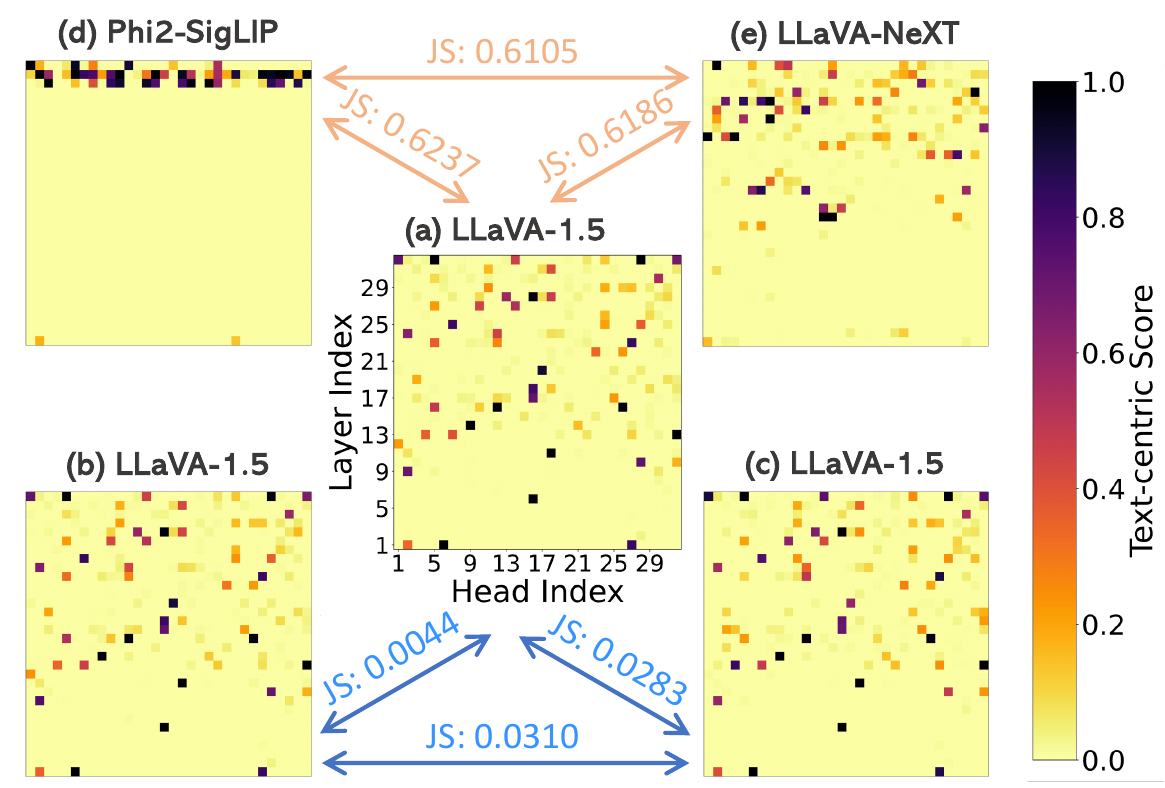}

   \caption{The stability of text-centric head distribution.
    Each heatmap visualizes how frequently a given head occurs among the most text-focused heads. LLaVA-1.5(a) remains stable across \emph{generation length}(b) and \emph{image set}(c), whereas Phi2-SigLIP(d) and LLaVA-NeXT(e) shift markedly.}
   \label{fig:mopic-head}
\end{figure}

In this section, we present our \emph{attention-steerable contrastive decoding} framework, which explicitly modulates the model’s attention to mitigate hallucinations. Our approach has two stages: (1)~\emph{Text-centric Head Selection}, which identifies the heads most prone to text-centric bias, and (2)~\emph{Attention Steering}, where we apply positive steering to text-centric heads and negative steering to a small subset of visually critical tokens. We then integrate these adjusted logits for generation into a contrastive decoding pipeline.

\subsection{Text-centric Heads are Model-specific}
\label{subsec:text-centric-heads}

Having established the impact of attention adjustments, we now discuss \emph{which} heads in the model are most prone to over-reliance on textual cues.
Here, we conduct an experiment to identify "text-centric" heads, i.e., those with disproportionately high text-to-visual attention ratios, and examine their consistency under different generation conditions and image sets. The experimental setup is detailed in the supplementary material (Appendix, Text-Centric Heads Experiment Settings).

\textbf{Results and Observations.}
Figure~\ref{fig:mopic-head} shows the resulting heatmaps $F$ for multiple models and generation settings. The panel in the center (a) corresponds to LLaVA-1.5 on $N=500$ images with a generation length of 64 tokens. The two heatmaps at the bottom show results of the same model but with either an increased generation length to 512 tokens (b, bottom left) or using a different set of 500 images (c, bottom right). Despite these changes, the distribution of top text-focused heads remains visually similar, and the small Jensen–Shannon (JS) divergences confirm that these text-centric heads are largely invariant under different sampling conditions for \emph{the same model}.

In contrast, the Phi2-SigLIP (d, top-left) and LLaVA-NeXT (e, top-right) panels deviate significantly from LLaVA-1.5 even under the same experimental settings, with higher JS divergence. This suggests that each model has its own unique set of heads that consistently favor textual attention over visual cues. However, \emph{within} a single model, the text-centric heads persist across varied prompts, image sets, and generation lengths.

\textbf{Implications.}
The consistent presence of the text-centric heads within the same model indicates that certain heads are inherently prone to focusing on textual signals rather than visual content. In the next subsection, we describe how this insight can be leveraged to selectively target the problematic heads when applying our \emph{positive steering} strategy.

\textbf{Robustness across Data Domains.}
To further assess the robustness of \emph{model-specific} text-centric heads, we repeat the profiling on extremely out-of-domain X-ray data. 
Details are provided in the supplementary material (“Extended Analysis of Text-centric Heads on Medical Data Domain”). 
The resulting heatmaps and Jensen–Shannon divergences show that, within each model, the \emph{same subset of heads} remains text-centric despite the shift from COCO photographs to chest X-rays, indicating strong domain robustness.

\begin{figure}[t]
  \centering
   \includegraphics[width=1.0\linewidth]{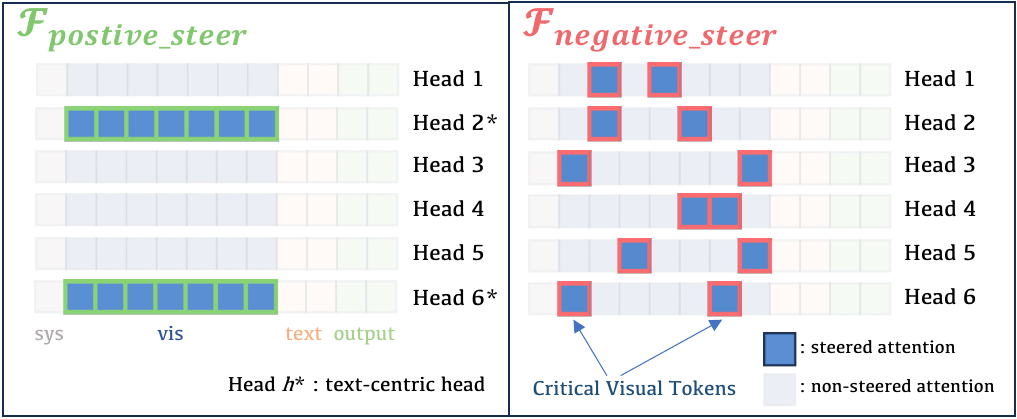}
   \caption{Illustration of positive and negative steering.
Left: text-centric heads are boosted (\emph{positive\_steer}) to emphasize visual content; Right: a small set of critical visual tokens is suppressed (\emph{negative\_steer}), inducing a stronger contrastive effect. These selective adjustments work in tandem to reduce hallucinations and improve grounding.}
   \label{fig:method-1}
\end{figure}

\subsection{Text-centric Head Selection}
\label{subsec:text-centric-head-selection}

\begin{algorithm}[t]
\caption{Text-Centric Head Selection (Offline)}
\label{alg:myopic-head-selection}

\KwIn{Reference image set $\{\mathbf{I}_1,\dots,\mathbf{I}_N\}$; MLLM with $L$ layers and $H$ heads per layer; desired text-centric head count $\kappa_{\textsc{tch}}$}
\KwOut{$\mathcal{H}_{\textsc{pos}}$ \tcp*[f]{Selected text-centric heads}}

\textbf{Initialize Global Statistics:}\\
Initialize counter tensor $F \leftarrow \mathbf{0}^{L\times H}$ \tcp*[f]{Vote accumulator}\\[2pt]

\textbf{Vote Accumulation Over Reference Set:}\\
\ForEach(\tcp*[f]{Iterate over images}){$\mathbf{I}_i \in \{\mathbf{I}_1,\dots,\mathbf{I}_N\}$}{
    Run MLLM on $\mathbf{I}_i$ to obtain cross-modal attentions\\
    \ForEach(\tcp*[f]{Loop over all heads}){$(r,c)\in\{1,\dots,L\}\times\{1,\dots,H\}$}{
        $Q_i(r,c)\leftarrow \dfrac{\text{textAttn}(r,c)}{\text{visAttn}(r,c)}$ \\
    }
    $\mathcal{I}_i \leftarrow$ indices of the \textbf{top-32} $Q_i$ values\\
    \ForEach{$(r,c)\in\mathcal{I}_i$}{
        $F(r,c)\leftarrow F(r,c)+1$ 
    }
}

\textbf{Head Selection:}\\
Sort heads $(r,c)$ by $F(r,c)$ in descending order\\
$\mathcal{H}_{\textsc{pos}}\leftarrow$ top $\kappa_{\textsc{tch}}$ heads\\

\Return $\mathcal{H}_{\textsc{pos}}$
\end{algorithm}

As detailed in Algorithm~\ref{alg:myopic-head-selection}, we start by identifying the most \emph{text-centric} heads using a small reference dataset (e.g., 500 images) for a task (e.g., image description). For each sample, we compute the ratio of textual attention to visual attention and take the top 32 heads with the highest ratio. We accumulate these counts over all samples, then choose the top $\kappa_{\textsc{tch}}$ heads as ``text-centric''. This step is motivated by our previous finding that certain heads consistently favor textual content over visual cues.

\begin{algorithm}[t]
\caption{Attention-Steerable Contrastive Decoding (ASCD)}
\label{alg:ascd}

\KwIn{Image $\mathbf{I}$; text-centric heads $\mathcal{H}_{\textsc{pos}}$; critical visual-token count $\kappa_{\textsc{vis}}$; steer strengths $\alpha_{\textsc{pos}},\alpha_{\textsc{neg}}$; contrastive weight $\alpha$; truncation threshold $\beta$; MLLM with $L$ layers and $H$ heads}
\KwOut{$p_\theta^{\text{final}}$}

\textbf{1. Positive Steering Pass:}\\
\For{$l\gets1$ \KwTo $L$}{
    \For{$h\gets1$ \KwTo $H$}{
        Compute attention matrix $\mathbf{A}^{(l)}_{h}$\\
        \If{$(l,h)\in\mathcal{H}_{\textsc{pos}}$}{
            $\mathbf{A}^{(l)}_{h}\leftarrow\mathbf{A}^{(l)}_{h}+\alpha_{\textsc{pos}}\lvert\mathbf{A}^{(l)}_{h}\rvert$
        }
    }
    Normalize $\mathbf{A}^{(l)}$ and continue
}
Obtain logits $p_\theta^{\text{pos}}$\\[4pt]

\textbf{2. Negative Steering Pass:}\\
\For{$l\gets1$ \KwTo $L$}{
    \For{$h\gets1$ \KwTo $H$}{
        Compute attention matrix $\mathbf{A}^{(l)}_{h}$\\
        Identify top-$\kappa_{\textsc{vis}}$ critical visual tokens $\mathcal{V_{\text{crit}}}$\\
        \ForEach{$v\in\mathcal{V_{\text{crit}}}$}{
            $\mathbf{A}^{(l)}_{h}(v)\leftarrow\mathbf{A}^{(l)}_{h}(v)-\alpha_{\textsc{neg}}\lvert\mathbf{A}^{(l)}_{h}(v)\rvert$
        }
    }
    Normalize $\mathbf{A}^{(l)}$ and continue
}
Obtain logits $p_\theta^{\text{neg}}$\\[4pt]

\textbf{3. Contrastive Decoding and Truncation:}\\
$p_\theta^{\text{raw}}\leftarrow(1+\alpha)\,p_\theta^{\text{pos}}-\alpha\,p_\theta^{\text{neg}}$\\
$\text{cutoff}\leftarrow\log(\beta)+\max(p_\theta^{\text{raw}})$\\
$p_\theta^{\text{final}}\leftarrow p_\theta^{\text{raw}}.\text{masked\_fill}(p_\theta^{\text{pos}}<\text{cutoff},-\infty)$\\
\Return $p_\theta^{\text{final}}$
\end{algorithm}

\subsection{Attention Steering}
\label{subsec:attention-steering}

\begin{figure}[t]
  \centering
   \includegraphics[width=1.0\linewidth]{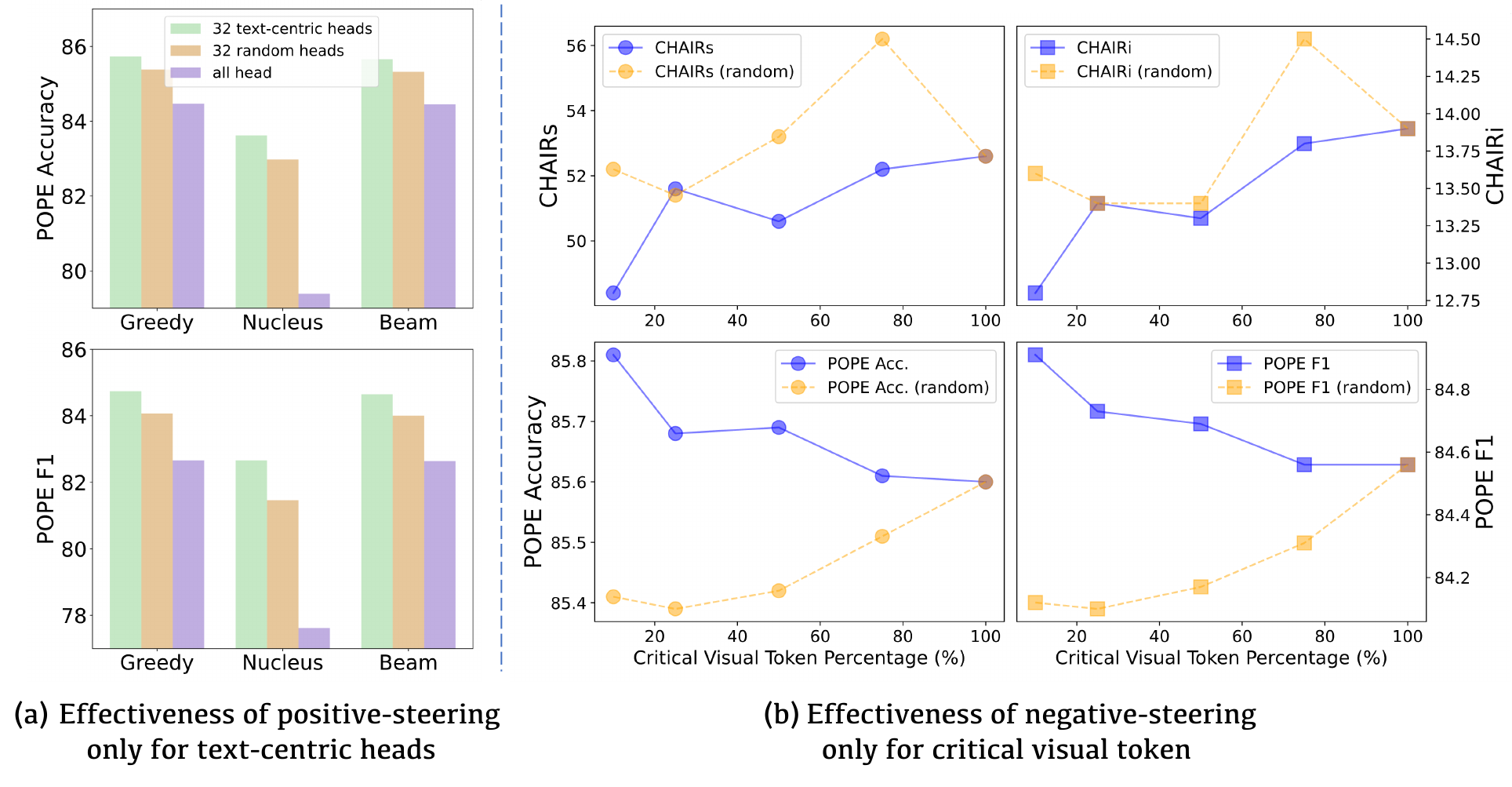}
    \caption{Comparative effectiveness of selective attention steering.    
    (a):~Positive steering applied \emph{only} to text-centric heads outperforms random or blanket head selection across various decoding strategies.
    (b):~Negative steering focused on a small subset of critical visual tokens, integrated with contrastive decoding, significantly reduces CHAIR metrics (less hallucination) and boosts POPE scores compared to randomly suppressing visual tokens of the same number.
    }
   \label{fig:method-2}
\end{figure}

\textbf{Text-centric Head Awareness and Critical Visual Token Selection.}  
As shown in Figure~\ref{fig:method-1}, we refine our method by incorporating text-centric head selection for positive steering and critical token identification for negative steering. Specifically, given the selected text-centric heads, we \emph{positively steer} them by increasing their attention weights with a strength of $\alpha_{\textsc{pos}}$. Figure \ref{fig:method-2}a highlights how targeted steering in text-centric heads improves the positive steering effectiveness. Simultaneously, we perform \emph{negative steering} on the $\kappa_{\textsc{vis}}$ visual tokens that draw the most attention—those with the highest head-averaged score.  
For the current query position, we define
\begin{equation}
    s(v)=\frac{1}{H}\sum_{h=1}^{H} A^{(L)}_{h}(v), \qquad v\in\{1,\dots,V\}.
    \label{eq:vis_score}
\end{equation}
The $\kappa_{\textsc{vis}}$ tokens with the largest $s(v)$ values form the critical set $\mathcal{V}_{\text{crit}}$; every attention entry to any $v\in\mathcal{V}_{\text{crit}}$ is then down-scaled in \emph{all} heads by $\alpha_{\textsc{neg}}\lvert A^{(L)}_{h}(v)\rvert$.

Through this strategy, we deliberately obscure only the most pivotal cues -- this targeted suppression is sufficient to induce a strong hallucination effect in the negative branch, leading to improved contrastive decoding compared to a blanket suppression of all visual tokens. In Figure \ref{fig:method-2}b, we demonstrate the impact of selectively applying negative steering to critical visual tokens.

\textbf{Integration with Contrastive Decoding with Truncation.}  
We first compute two output distributions: \(p_{\theta}^{\text{pos}}\) from the positively steered branch and \(p_{\theta}^{\text{neg}}\) from the negatively steered branch.  
Step~3 of Algorithm~\ref{alg:ascd} then fuses them through a contrastive-decoding rule with truncation, producing the final logits.  
This fusion amplifies visually grounded evidence while suppressing spurious text-only cues, thereby reducing hallucinations.

\section{Experiments}
\label{sec:expeirments}

\begin{table*}
    \centering
    \setlength{\tabcolsep}{4mm}
    \small
    \begin{tabular}{ccccccc}
        \toprule
        Model & Decoding & Method & CHAIRs (↓) & CHAIRi (↓) & POPE-Acc (↑)& POPE-F1 (↑) \\
        \midrule
        \multirow{15}{*}{\makecell{LLaVA-\\1.5 7B}} 
        & \multirow{5}{*}{greedy} & Orig & 53.2 & 13.5 & 85.37 & 84.06 \\
        & & VCD  & 56.8 & 15.2 & 84.27 & 83.35 \\
        & & ICD  & 52.8 & 13.2 & 83.07 & 80.64 \\
        & & PAI  & - & - & 85.82 & 85.79 \\

         & & ASCD  & \textbf{35.6 (33.1\%)} & \textbf{8.6 (36.3\%)} & \textbf{86.53} & \textbf{86.25} \\
        \cmidrule(lr){2-7}
        & \multirow{5}{*}{nucleus} & Orig & 59.0 & 17.4 & 83.03 & 81.57 \\
        & & VCD  & 59.8 & 16.6  & 83.31 & 82.30 \\
        & & ICD  & 57.4 & 15.6 & 82.13 & 79.62 \\
        & & PAI & - & - & 81.72 & 82.87 \\

         & & ASCD  & \textbf{43.6 (26.1\%)} & \textbf{11.3 (35.1\%)} & \textbf{85.75} & \textbf{85.07} \\
        \cmidrule(lr){2-7}
        & \multirow{5}{*}{beam} & Orig & 54.8 & 15.3 & 85.40 & 84.10 \\
        & & VCD  & 58.8 & 16.4 & 84.27 & 83.30 \\
        & & ICD  & 52.6 & 13.9 & 83.04 & 80.59  \\
        & & PAI & -  & - & 86.33 & 85.89 \\

         & & ASCD & \textbf{40.8 (25.5\%)} & \textbf{10.1 (34.0\%)} & \textbf{86.52} & \textbf{86.24}\\
        \midrule


        \multirow{12}{*}{\makecell{LLaVA-\\NeXT 7B}} 
        & \multirow{4}{*}{greedy} & Orig & 31.6 & 7.5 & 83.93 & 81.89  \\
        & & VCD  & 37.2 & 9.7  & \textbf{84.86} & 83.28 \\
        & & ICD  & 32.8 & 8.4 & 84.44 & 82.70 \\
         & & ASCD  & \textbf{21.8 (31.0\%)} & \textbf{7.0 (6.7\%)} & \underline{84.85} & \textbf{83.40} \\
        \cmidrule(lr){2-7}
        & \multirow{4}{*}{nucleus} & Orig & 30.4 & 8.0  & 81.74 & 79.61 \\
        & & VCD  & 40.4 & 10.4 & 83.55 & 81.95 \\
        & & ICD  & 39.4 & 9.9 & 83.67 & 81.98 \\
          & & ASCD & \textbf{21.2 (30.3\%)} & \textbf{6.7 (16.3\%)} & \textbf{84.69} & \textbf{83.09} \\
        \cmidrule(lr){2-7}
        & \multirow{4}{*}{beam} & Orig & 34.0 & 8.5& 84.11 & 82.14 \\
        & & VCD  & 36.6 & 9.1& 84.66 & 83.03 \\
        & & ICD  & 31.8 & 7.6& 84.48 & 82.75  \\
         & & ASCD  & \textbf{21.0 (38.2\%)} & \textbf{6.5 (23.5\%)} & \textbf{84.91} & \textbf{83.48}\\
        \midrule
        \multirow{12}{*}{\makecell{Phi2-\\SigLIP}} 
        & \multirow{4}{*}{greedy} & Orig & 29.0 & 6.9 & 87.19 & 86.16 \\
        & & VCD  & 39.4 & 9.6 & 86.22 & 85.53 \\
        & & ICD  & 33.4 & 7.7  & 85.83 & 84.58\\
         & & ASCD  & \textbf{21.8 (24.8\%)} & \textbf{5.4 (21.7\%)} & \textbf{87.81} & \textbf{86.90} \\
        \cmidrule(lr){2-7}
        & \multirow{4}{*}{nucleus} & Orig & 36.0 & 9.8 & 85.51 & 84.44\\
        & & VCD  & 36.0 & 8.1 & 85.60 & 84.86 \\
        & & ICD  & 37.0 & 9.4 & 84.63 & 83.35 \\
         & & ASCD  & \textbf{26.0 (27.8\%)} & \textbf{8.0 (18.4\%)}  & \textbf{87.45} & \textbf{86.46}\\
        \cmidrule(lr){2-7}
        & \multirow{4}{*}{beam} & Orig & 30.4 & 6.9 & 87.19 & 86.16 \\
        & & VCD  & 36.0 & 8.4 & 86.30 & 85.64  \\
        & & ICD  & 31.0 & 7.0 & 85.83 & 84.58\\
        
         & & ASCD   & \textbf{24.6 (19.1\%)} & \textbf{5.7 (17.4\%)} & \textbf{87.81} & \textbf{86.90} \\

    \midrule
        \multirow{4.5}{*}{\makecell{LLaVA-\\1.5 13B}} 
        & \multirow{2}{*}{greedy} & Orig &51.2 & 12.6 & 85.52 & 84.12 \\
         & & ASCD  & \textbf{33.0(35.5\%)} & \textbf{8.0(36.5\%)} & \textbf{87.78} & \textbf{87.40} \\
        \cmidrule(lr){2-7}
        & \multirow{2}{*}{nucleus} & Orig &51.4 & 14.5& 83.95 & 82.54\\
         & & ASCD  & \textbf{35.7(30.5\%)} & \textbf{9.5(34.5\%)}  & \textbf{87.19} & \textbf{86.46}\\

    \midrule
        \multirow{4.5}{*}{\makecell{Qwen-\\2.5-VL-\\Instruct}} 
        & \multirow{2}{*}{greedy} & Orig &31.2 & 7.5 & 87.72 & 86.64 \\
         & & ASCD  & \textbf{24.4(21.8\%)} & \textbf{6.4(14.7\%)} & \textbf{88.91} & \textbf{88.27} \\
        \cmidrule(lr){2-7}
        & \multirow{2}{*}{nucleus} & Orig &34.8 & 8.6& 87.34 & 86.14\\
         & & ASCD  & \textbf{26.9(22.7\%)} & \textbf{6.3(26.7\%)}  & \textbf{88.45} & \textbf{87.68}\\

        \bottomrule
    \end{tabular}
    \caption{CHAIR and POPE Evaluation Results. Lower CHAIRs and CHAIRi values indicate better performance in reducing hallucinations. POPE performance is reported as the mean accuracy and F1 score. The best values for each metric within a model-decoding combination are highlighted in bold. If \mbox{ASCD} ranks second, the best is bold while the \mbox{ASCD} score is \underline{under\-lined}.}
    \label{tab:chair_pope_results}
\end{table*}

To evaluate the effectiveness of our attention-steerable contrastive decoding framework in mitigating hallucinations in MLLMs, we conduct a range of experiments. This includes three diverse benchmarks—\textbf{CHAIR}, \textbf{POPE}, and \textbf{MMHal-Bench}—each designed to assess different aspects of object hallucinations. To ensure the broad applicability and robustness of our approach, we test it on three representative models—\textbf{LLaVA-1.5 7B}, \textbf{LLaVA-NeXT 7B}, \textbf{Phi2-SigLIP},  and employ three different decoding strategies: \textbf{greedy search}, \textbf{nucleus sampling}, and \textbf{beam search}. Details of the experimental settings are provided in the supplementary material (Appendix, Evaluation Settings).
Furthermore, we evaluate performance on standard VQA benchmarks including \textbf{MMMU}, \textbf{MM-VET}, \textbf{ScienceQA}, \textbf{TextVQA}, and \textbf{GQA} to verify that the proposed method preserves—rather than diminishes—the model’s original visual understanding.
In addition, an \emph{extended} evaluation on the larger \textbf{LLaVA-1.5 13B} and the modern \textbf{Qwen2.5-VL-Instruct} is conducted to verify scalability and architectural generality on a subset of benchmarks.





\begin{figure}[b]
  \centering
   \includegraphics[width=1.0\linewidth]{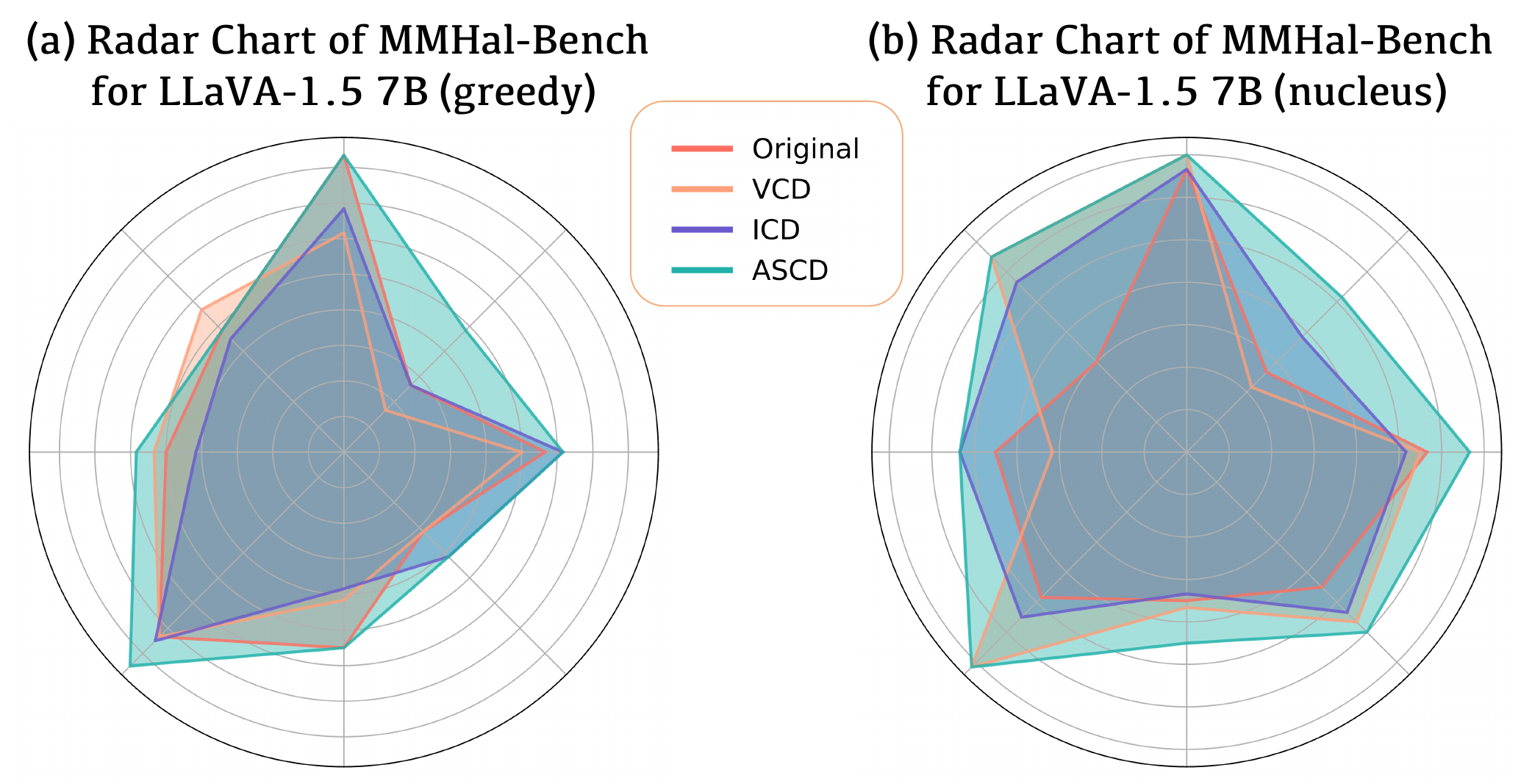}

   \caption{Radar charts of MMHal-Bench results. Each axis represents a different evaluation dimension in MMHal-Bench, and a larger enclosed area indicates better overall performance.}
   \label{fig:mmhal-radar}
\end{figure}








It is important to note that current benchmarks for evaluating multimodal models are highly variable. For example, baseline models such as LLaVA-1.5 7B often report different metric values between different papers. Moreover, the CHAIR metric relies on random image sampling, which further complicates direct comparisons between papers. To address these issues, we faithfully \emph{reproduced} both VCD and ICD using the parameters specified in their original papers and repositories, ensuring that our evaluations are conducted under consistent conditions.

\textbf{POPE and CHAIR.}
Table~\ref{tab:chair_pope_results} summarizes both caption-level (\textsc{CHAIR}) and VQA-style (\textsc{POPE}) results.  
Across every backbone and decoding scheme, ASCD produces the \emph{lowest} CHAIR scores and the \emph{highest} POPE accuracy/F1, outperforming Orig, VCD, and ICD. These gains hold for all three prompt types (random, popular, adversarial; see Appendix~“Detailed POPE Results”) and remain stable when scaling from 7 B to 13 B or switching to the Qwen-VL architecture, indicating that attention steering mitigates object-level hallucinations in a model- and domain-robust manner.

\textbf{MMHal-Bench.}
Figure~\ref{fig:mmhal-radar} illustrates the radar charts of MMHal-Bench results for LLaVA-1.5 7B under greedy and nucleus decoding. Each axis represents a sub-dimension of the benchmark, and a larger area signifies better overall performance. ASCD exhibits the largest enclosed area, outperforming baseline, VCD, and ICD in most dimensions.

\textbf{Standard VQA Benchmarks.}
To verify that ASCD does not sacrifice a model’s general visual-question-answering ability, it's evaluated on five widely-used VQA datasets.
Across all three representative backbones and all decoding strategies, ASCD either matches or surpasses the original model on every dataset, while VCD and ICD consistently degrade performance as shown in Table~\ref{tab:vqa_scores}.

\textbf{Summary.}
Our experiments confirm that ASCD effectively reduces hallucinations and improves alignment with visual content, regardless of the model or decoding strategy employed.

\textbf{Additional Experiments.}
The supplementary material presents a runtime/memory profile and a comprehensive hyper-parameter ablation.
Together, these studies confirm that ASCD adds only marginal computational overhead and remains effective across a broad range of settings, making it practical to deploy with minimal tuning effort.

\begin{table}[t]
    \centering
    
    \begin{tabular}{lcccc}
        \toprule
        \textbf{Benchmark} & Orig & VCD & ICD & \textbf{ASCD} \\ 
        \midrule
        MM-VET (↑)    & 31.2   & 30.3   & 33.2   & \textbf{33.2} \\
        ScienceQA (↑) & 67.55  & 67.55  & 67.32  & \textbf{69.51} \\
        GQA (↑)       & \textbf{61.28} & 59.38 & 59.99 & \underline{61.27} \\
        TextVQA (↑)   & 57.82  & 55.07  & 57.66  & \textbf{57.91} \\
        MMMU (↑)      & 0.342  & 0.333  & \textbf{0.360} & \underline{0.348} \\
        \bottomrule
    \end{tabular}
    \caption{Scores on five VQA-style benchmarks. The best score in each row is bold.  
    If \mbox{ASCD} ranks second, the best is bold while the \mbox{ASCD} score is \underline{under\-lined}.}
    \label{tab:vqa_scores}
\end{table}
\section{Conclusion}
\label{sec:conclusion}

We have shown that existing contrastive methods (e.g., VCD and ICD) inadvertently \emph{shift} the internal attention distribution in MLLMs, prompting us to investigate a more direct and principled way to modulate attention. We propose an \emph{attention-steerable contrastive decoding} framework that \emph{positively steers} text-centric heads while \emph{negatively steering} only the most critical visual tokens.


Our method consistently reduces hallucinations on CHAIR, POPE, and MMHal-Bench, outperforming both baseline and previous contrastive approaches with improved and uncompromised general VQA capability. By targeting precisely those heads and tokens, we effectively mitigate spurious textual biases while preserving essential visual context.

\bibliography{main}
\appendix
\section*{Appendix} 
\section{Text-centric Heads Experiment Settings}
\label{sec:text_centric_heads_exp_settings}

The following setup applies to "Subsection: Text-centric Heads are Model-specific".

We select $N=500$ images $\{\mathbf{I}_1,\dots,\mathbf{I}_N\}$ (from COCO) and run an MLLM (LLaVA-1.5) in an image description task. During each generation, we track the ratio of textual attention to visual attention for every head:
\begin{equation}
    Q_i \in \mathbb{R}^{R \times C}, 
    \quad
    Q_i(r, c) \;=\;
    \frac{\text{textAttn}(r,c)}{\text{visAttn}(r,c)},
    \label{eq:attn-ratio}
\end{equation}
where $r$ and $c$ index each head (for instance, $R=C=32$). We then identify the top-$k$ heads with the highest ratio values and mark them in a binary mask:
\begin{equation}
    M_i(r,c)
    \;=\;
    \begin{cases}
        1, & \text{if }(r,c)\in \text{top-$k$ indices of } Q_i,\\
        0, & \text{otherwise.}
    \end{cases}
    \label{eq:mask}
\end{equation}
Finally, we aggregate these masks across all $N$ images:
\begin{equation}
    F \;=\; \sum_{i=1}^{N} M_i,
    \label{eq:frequency}
\end{equation}
so that $F(r,c)$ records how frequently head $(r,c)$ appears among the most text-focused heads.

\section{Extended Analysis of Text-centric Heads on Medical Data Domain}
\label{sec:extended_analysis_text_centric_heads}

\paragraph{Goal.}
The heads identified as text-centric were profiled on a COCO reference set containing natural photographs. This is already a very general and ideal setting for selecting text-centric heads.
However, we would like to verify that this selection is not tied to the training domain but even in an extremely OOD setting, we repeat the profiling procedure on a visually dissimilar corpus of 500 NIH ChestX-ray14 images.

\paragraph{Protocol.}
For every backbone we rerun the counting algorithm on the X-ray set, obtain a new head-frequency tensor $F_{\text{X-ray}}$, and compute the Jensen–Shannon divergence (JSD) between the COCO and X-ray distributions.  
For comparison we also report \emph{inter-model} JSD values within each domain.

\begin{table}[t]
\centering
\setlength{\tabcolsep}{4pt}  

\begin{tabular}{lccc}
\toprule
\multicolumn{4}{c}{\textbf{Intra-model JS Divergence (COCO vs.\ X-ray)}}\\
\midrule
& LLaVA-1.5 & LLaVA-NeXT & Phi-2-SigLIP \\
\cmidrule(lr){2-4}
JSD & 0.101 & 0.094 & 0.092 \\
\bottomrule
\end{tabular}

\begin{tabular}{lccc}
\toprule
\multicolumn{4}{c}{\textbf{Inter-model JS Divergence on COCO}}\\
\midrule
& L-1.5 vs.\ L-NeXT & L-NeXT vs.\ P-S & L-1.5 vs.\ P-S \\
\cmidrule(lr){2-4}
JSD & 0.619 & 0.611 & 0.624 \\
\bottomrule
\end{tabular}

\begin{tabular}{lccc}
\toprule
\multicolumn{4}{c}{\textbf{Inter-model JS Divergence on X-ray}}\\
\midrule
& L-1.5 vs.\ L-NeXT & L-NeXT vs.\ P-S & L-1.5 vs.\ P-S \\
\cmidrule(lr){2-4}
JSD & 0.604 & 0.572 & 0.620 \\
\bottomrule
\end{tabular}

\caption{Jensen–Shannon divergence between text-centric-head distributions.
A value below 0.1 is widely regarded as negligible drift.
Abbreviations: L = LLaVA, P-S = Phi-2-SigLIP.}
\label{tab:jsd_threeblocks}
\end{table}

\paragraph{Result.}
As shown in Table~\ref{tab:jsd_threeblocks}, all three backbones exhibit intra-model JSD $\le 0.11$,
signalling that the same set of heads remains text-centric
even when the reference data shift significantly from COCO to chest X-rays.
By contrast, swapping architectures while keeping the domain fixed yields JSD ${\sim}0.60$, confirming that the identity of text-centric heads is governed by model internals rather than by the reference dataset.

\section{Extended Evaluation: Model Scaling and Modern Architecture}
\label{sec:extended_eval_on_scal_model_and_modern_architecture}

\subsection{Further Evaluation}
\label{subsec:further_eval}

We further benchmark ASCD on (i) \textbf{LLaVA-1.5-13B}, to test \emph{scaling} effects, and  
(ii) \textbf{Qwen2.5-VL-Instruct}, one of the latest open‐source vision–language models.  
Table \ref{tab:pope_extended} reports POPE accuracy/F1 under greedy and nucleus decoding.  
Across both backbones and decoding schemes, ASCD yields consistent gains, demonstrating that our attention-steering approach scales beyond 3B and 7B models and transfers across different architectures.

\begin{table*}[t]
\centering
\setlength{\tabcolsep}{1mm}
\begin{tabular}{ccccccccccc}
\toprule
        \multirow{2}{*}{Model} & \multirow{2}{*}{Dec.} & \multirow{2}{*}{Method} &
        \multicolumn{2}{c}{Popular} & \multicolumn{2}{c}{Random} &
        \multicolumn{2}{c}{Adversarial} & \multicolumn{2}{c}{Average} \\
        \cmidrule(lr){4-11}
         & & &
        \makecell{Acc} & \makecell{F1} &
        \makecell{Acc} & \makecell{F1} &
        \makecell{Acc} & \makecell{F1} &
        \makecell{Acc} & \makecell{F1} \\
\midrule
\multirow{4}{*}{\makecell{LLaVA-\\1.5 13B}}
    & \multirow{2}{*}{greedy}
        & Orig & 0.860 & 0.844 & 0.866 & 0.853 & 0.840 & 0.826 & 0.855 & 0.841 \\

    &   & ASCD & \textbf{0.883} & \textbf{0.878} & \textbf{0.899} & \textbf{0.896} & \textbf{0.848} & \textbf{0.847} & \textbf{0.877} & \textbf{0.874} \\
    & \multirow{2}{*}{nucleus}
        & Orig & 0.842 & 0.827 & 0.848 & 0.835 & 0.826 & 0.814 & 0.839 & 0.825 \\

    &   & ASCD & \textbf{0.876} & \textbf{0.867} & \textbf{0.887} & \textbf{0.880} & \textbf{0.851} & \textbf{0.845} & \textbf{0.871} & \textbf{0.864} \\
\midrule
\multirow{4}{*}{\makecell{Qwen2.5-\\VL-Instruct}}
    & \multirow{2}{*}{greedy}
        & Orig & 0.878 & 0.865 & 0.886 & 0.876 & 0.868 & 0.857 & 0.877 & 0.866 \\

    &   & ASCD & \textbf{0.891} & \textbf{0.883} & \textbf{0.902} & \textbf{0.896} & \textbf{0.875} & \textbf{0.868} & \textbf{0.889} & \textbf{0.882} \\
    & \multirow{2}{*}{nucleus}
        & Orig & 0.876 & 0.864 & 0.880 & 0.870 & 0.862 & 0.850 & 0.873 & 0.861 \\

    &   & ASCD & \textbf{0.889} & \textbf{0.881} & \textbf{0.895} & \textbf{0.888} & \textbf{0.869} & \textbf{0.860} & \textbf{0.884} & \textbf{0.876} \\
\bottomrule
\end{tabular}
\caption{POPE benchmark results on a \emph{larger} (LLaVA-1.5 13B) and a \emph{novel} (Qwen2.5-VL-Instruct) backbone.  Bold numbers mark improvements of ASCD over the corresponding original model under the same decoding strategy; gray rows highlight ASCD.}
\label{tab:pope_extended}
\end{table*}

\subsection{Visualization of Text-centric Heads Distribution of LLaVA-1.5 13B and Qwen2.5-VL-Instruct}
\label{subsec:vis_tch_distribution}

\begin{figure}
  \centering
   \includegraphics[width=1.0\linewidth]{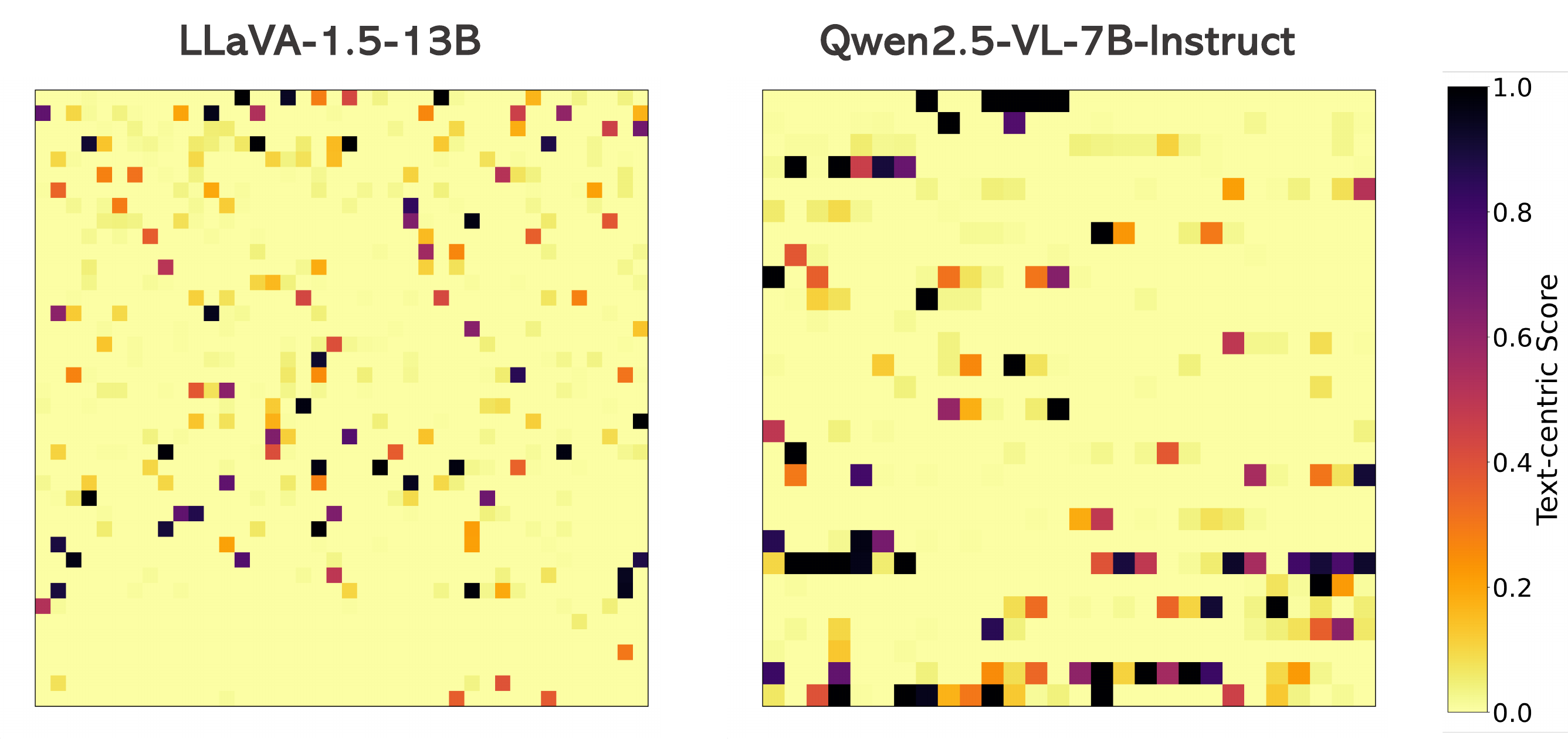}

   \caption{Text-centric-head frequency maps.
  Each pixel denotes the probability that a given layer–head is selected as
  text-centric when profiled on 500 COCO images.
  Darker colours highlight a handful of heads that dominate textual attention,
  revealing the same long-tailed, head-level anisotropy seen in previous models.}
   \label{fig:qwen_llava_13b_heatmap}
\end{figure}

Figure \ref{fig:qwen_llava_13b_heatmap} visualizes the head-frequency heatmap for
\textbf{LLaVA-1.5 13B} and \textbf{Qwen2.5-VL-Instruct}.
Both models reproduce the \emph{sparse, high-contrast pattern} already observed for the previous backbones:
a small set of heads appear in dark colors, indicating that they are marked
text-centric in the vast majority of reference samples,
while the remainder receive little or no votes.

\section{Detailed POPE Results}
\label{sec:detailed_pope_results}

Table~\ref{tab:detailed_pope_results} presents the full POPE breakdown by
different type—\emph{random}, \emph{popular}, and \emph{adversarial}—for each
backbone and decoding scheme. We report accuracy and F1.
Across all settings, \textbf{ASCD} consistently matches or surpasses the
baselines, with the largest gains typically appearing on the random split.

\begin{table*}[t!]
    \centering
    \begin{tabular}{ccccccccccc}
        \toprule
        \multirow{2}{*}{Model} & \multirow{2}{*}{Dec.} & \multirow{2}{*}{Method} &
        \multicolumn{2}{c}{Popular} & \multicolumn{2}{c}{Random} &
        \multicolumn{2}{c}{Adversarial} & \multicolumn{2}{c}{Average} \\
        \cmidrule(lr){4-11}
         & & &
        \makecell{Acc} & \makecell{F1} &
        \makecell{Acc} & \makecell{F1} &
        \makecell{Acc} & \makecell{F1} &
        \makecell{Acc} & \makecell{F1} \\
        \midrule
        \multirow{15}{*}{LLaVA-1.5 7B} 
        & \multirow{4}{*}{greedy} & Orig & 85.83 & 84.35 & 86.67 & 85.51 & \textbf{83.60} & 82.31 & 85.37 & 84.06 \\
        & & VCD  & 84.67 & 83.59 & 86.43 & 85.53 & 81.70 & 80.93 & 84.27 & 83.35 \\
        & & ICD  & 83.47 & 80.85 & 83.68 & 81.51 & 82.07 & 79.56 & 83.07 & 80.64 \\
        & & PAI & - & - & - & - & - & - & 85.82 & 85.79 \\
         & & ASCD  & \textbf{87.20} & \textbf{86.69} & \textbf{89.48} & \textbf{89.09} & \underline{82.90} & \textbf{82.97} & \textbf{86.53} & \textbf{86.25} \\
        \cmidrule(lr){2-11}
        & \multirow{4}{*}{nucleus} & Orig & 83.47 & 81.74 & 84.78 & 83.44 & 80.83 & 79.53 & 83.03 & 81.57 \\
        & & VCD  & 83.83 & 82.61 & 85.12 & 84.12 & 80.97 & 80.18 & 83.31 & 82.30 \\
        & & ICD  & 82.63 & 79.85 & 82.68 & 80.43 & 81.07 & 78.58 & 82.13 & 79.62 \\
        & & PAI & - & - & - & - & - & - & 81.72 & 82.87 \\
         & & ASCD  & \textbf{86.47} & \textbf{85.56} & \textbf{87.94} & \textbf{87.26} & \textbf{82.83} & \textbf{82.39} & \textbf{85.75} & \textbf{85.07} \\
        \cmidrule(lr){2-11}
        & \multirow{4}{*}{beam} & Orig & 85.87 & 84.39 & 86.70 & 85.55 & \textbf{83.63} & 82.36 & 85.40 & 84.10 \\
        & & VCD  & 84.43 & 83.30 & 86.19 & 85.31 & 82.20 & 81.30 & 84.27 & 83.30 \\
        & & ICD  & 83.47 & 80.83 & 83.61 & 81.43 & 82.03 & 79.51 & 83.04 & 80.59 \\
        & & PAI & - & - & - & - & - & - & 86.33 & 85.89 \\
         & &  ASCD  & \textbf{87.20} & \textbf{86.69} & \textbf{89.48} & \textbf{89.09} & \underline{82.87} & \textbf{82.95} & \textbf{86.52} & \textbf{86.24} \\
        \midrule
        \multirow{12}{*}{LLaVA-NeXT 7B} 
        & \multirow{4}{*}{greedy} & Orig & 83.97 & 81.77 & 85.09 & 83.26 & 82.73 & 80.64 & 83.93 & 81.89 \\
        & & VCD  & 84.87 & 83.12 & 86.19 & 84.84 & \textbf{83.53} & \textbf{81.89} & \textbf{84.86} & 83.28 \\
        & & ICD  & 84.53 & 82.63 & 85.70 & 84.15 & 83.10 & 81.33 & 84.44 & 82.70 \\
         & & ASCD & \textbf{84.90} & \textbf{83.30} & \textbf{86.39} & \textbf{85.09} & \underline{83.27} & \underline{81.82} & \underline{84.85} & \textbf{83.40} \\
        \cmidrule(lr){2-11}
        & \multirow{4}{*}{nucleus} & Orig & 81.73 & 79.26 & 83.61 & 81.75 & 79.87 & 77.81 & 81.74 & 79.61 \\
        & & VCD  & 84.20 & 82.51 & 84.78 & 83.26 & 81.67 & 80.07 & 83.55 & 81.95 \\
        & & ICD  & 83.60 & 81.68 & 85.29 & 83.78 & 82.13 & 80.47 & 83.67 & 81.98 \\
         & &  ASCD  & \textbf{84.60} & \textbf{82.86} & \textbf{86.19} & \textbf{84.77} & \textbf{83.27} & \textbf{81.65} & \textbf{84.69} & \textbf{83.09} \\
        \cmidrule(lr){2-11}
        & \multirow{4}{*}{beam} & Orig & 84.17 & 82.04 & 85.26 & 83.49 & 82.90 & 80.88 & 84.11 & 82.14 \\
        & & VCD  & 84.67 & 82.86 & 86.19 & 84.81 & 83.13 & 81.42 & 84.66 & 83.03 \\
        & & ICD  & 84.57 & 82.68 & 85.74 & 84.19 & 83.13 & 81.37 & 84.48 & 82.75 \\
         & &  ASCD  & \textbf{84.97} & \textbf{83.39} & \textbf{86.43} & \textbf{85.14} & \textbf{83.33} & \textbf{81.91} & \textbf{84.91} & \textbf{83.48} \\
        \midrule
        \multirow{12}{*}{Phi2-SigLIP} 
        & \multirow{4}{*}{greedy} & Orig & 87.10 & 85.95 & 88.45 & 87.57 & 86.03 & 84.97 & 87.19 & 86.16 \\
        & & VCD  & 86.00 & 85.14 & 87.97 & 87.37 & 84.70 & 84.09 & 86.22 & 85.53 \\
        & & ICD  & 85.50 & 84.14 & 87.25 & 86.15 & 84.73 & 83.44 & 85.83 & 84.58 \\
         & & ASCD  & \textbf{87.77} & \textbf{86.74} & \textbf{88.90} & \textbf{88.14} & \textbf{86.77} & \textbf{85.81} & \textbf{87.81} & \textbf{86.90} \\
        \cmidrule(lr){2-11}
        & \multirow{4}{*}{nucleus} & Orig & 85.73 & 84.49 & 86.87 & 85.87 & 83.93 & 82.96 & 85.51 & 84.44 \\
        & & VCD  & 85.60 & 84.72 & 86.91 & 86.21 & 84.30 & 83.64 & 85.60 & 84.86 \\
        & & ICD  & 84.90 & 83.46 & 85.98 & 84.86 & 83.00 & 81.73 & 84.63 & 83.35 \\
         & &   ASCD  & \textbf{87.50} & \textbf{86.41} & \textbf{88.52} & \textbf{87.69} & \textbf{86.33} & \textbf{85.29} & \textbf{87.45} & \textbf{86.46} \\
        \cmidrule(lr){2-11}
        & \multirow{4}{*}{beam} & Orig & 87.10 & 85.95 & 88.45 & 87.57 & 86.03 & 84.97 & 87.19 & 86.16 \\
        & & VCD  & 86.43 & 85.63 & 87.90 & 87.36 & 84.57 & 83.92 & 86.30 & 85.64 \\
        & & ICD  & 85.50 & 84.14 & 87.25 & 86.15 & 84.73 & 83.44 & 85.83 & 84.58 \\
         & & ASCD  & \textbf{87.77} & \textbf{86.74} & \textbf{88.90} & \textbf{88.14} & \textbf{86.77} & \textbf{85.81} & \textbf{87.81} & \textbf{86.90} \\
        \bottomrule
    \end{tabular}
    \caption{POPE Evaluation Results. The best values for each metric within a model-decoding combination are highlighted in bold. If our ASCD achieves the second-best result, it is additionally marked with an \underline{underline}.}
    \label{tab:detailed_pope_results}
\end{table*}

\section{Evaluation Settings}
\label{sec:exp_settings}

\subsection{Baseline Models and Decoding Methods}
\label{subsec:intro_for_models_decode}

We evaluate our proposed approach using three representative models and multiple decoding strategies to demonstrate its broad applicability and robustness.

\textbf{LLaVA-1.5 7B} is a minimalist yet efficient model that has served as the foundation for extensive studies in large multimodal modeling. \textbf{LLaVA-NeXT 7B} builds on LLaVA-1.5 with improvements in visual reasoning, higher input resolution, and enhanced world knowledge, resulting in superior performance on several benchmarks. \textbf{Phi2-SigLIP} leverages the Phi-2 backbone and a SigLIP-based vision tower, and is trained on the ShareGPT4V dataset, offering a compact alternative with competitive capabilities. \textbf{LLaVA-1.5 13B} upsizes LLaVA-1.5 to 13B parameters—doubling language capacity while retaining the original architecture—so it serves as a clean test bed for model-scaling effects. \textbf{Qwen2.5-VL-Instruct} pairs the recent Qwen2.5-7B language model with a revamped vision encoder and multi-stage instruction tuning, representing a modern open-source architecture that achieves state-of-the-art performance out of the box.

To assess the reliability and generalizability, we experiment with three decoding strategies: \textbf{greedy search}, \textbf{nucleus sampling}, and \textbf{beam search}. Greedy decoding yields deterministic outputs, while nucleus sampling and beam search enable more diverse generation.

\subsection{Datasets}
\label{subsec:intro_for_datasets}

With the widespread adoption of deep learning across computer vision and natural language processing, large language models (LLMs) have rapidly become foundational tools in AI systems. However, despite their remarkable capabilities, MLLMs remain susceptible to hallucination—generating descriptions inconsistent with the input image. This phenomenon has sparked growing interest in evaluation benchmarks that specifically target hallucination in image-grounded generation.

We evaluate our approach on three hallucination-targeted benchmark datasets designed to probe object hallucination in multimodal large language models.

\textbf{CHAIR.}
The Caption Hallucination Assessment with Image Relevance (CHAIR) metric quantifies the degree of hallucination in generated captions by measuring the fraction of objects mentioned that do not actually appear in the image. It is computed at both the instance-level (CHAIRi) and the sentence-level (CHAIRs), offering insight into how well a caption adheres to veridical image content. 

\textbf{POPE.}
The Polling-based Object Probing Evaluation (POPE) assesses hallucination by querying the model with binary questions (\eg, “Is there a car in the image?”). By balancing queries about present and absent objects, and using different sampling strategies (random, popular, adversarial), POPE effectively reveals the influence of language priors on model predictions. This method provides a robust measure of object hallucination across multiple datasets such as MSCOCO, A-OKVQA, and GQA.

\textbf{MMHal-Bench.}
MMHal-Bench is a new evaluation benchmark specifically designed to challenge large multimodal models in hallucination. Comprising 96 difficult questions based on images, along with detailed ground-truth answers and image content annotations, MMHal-Bench offers a comprehensive testbed to assess the model’s ability to produce veridical responses in ambiguous visual scenarios.

We also evaluate our approach on five standard VQA benchmarks to verify that our method does not compromise, and may even enhance, the original VQA capabilities.

\textbf{MM-VET.}
The MM-VET benchmark evaluates large multimodal models on integrated vision–language abilities that go beyond conventional VQA. It defines six core capabilities—recognition, OCR, knowledge, language generation, spatial awareness, and math—and constructs 16 task types that require their various combinations. MM-VET comprises 200 images paired with 218 open-ended questions; answers are graded by a GPT-4-based evaluator that yields unified scores, enabling fine-grained, per-capability diagnostics.

\textbf{ScienceQA.}
The ScienceQA benchmark targets multimodal, curriculum-level science reasoning. It contains multiple-choice questions drawn from elementary- to high-school materials, each optionally paired with text passages, natural or diagrammatic images, or both. In this study, answers are graded for accuracy.

\textbf{GQA.}
The GQA benchmark targets real-world compositional visual reasoning. Questions span objects, attributes, relations, spatial logic, comparisons, and multi-hop inference, making GQA a stringent yardstick for models that aspire to deep vision-language understanding rather than pattern-matching.

\textbf{TextVQA.}
The TextVQA benchmark focuses on visual question answering that explicitly requires reading and reasoning over scene text. It contains 45,336 open-ended questions paired with 28,408 everyday images drawn from the Open Images dataset. Questions span practical tasks such as reading product labels, traffic signs, receipts, and clocks, so models must combine OCR, spatial grounding, and commonsense reasoning.

\textbf{MMMU.}
The Massive Multi-discipline Multimodal Understanding and Reasoning (MMMU) benchmark gauges college-level, expert-style reasoning across six broad disciplines—Art \& Design, Business, Science, Health \& Medicine, Humanities \& Social Science, and Tech \& Engineering.

\subsection{Hyperparameters}
\label{subsec:hyperparameters}

The experiments are conducted with the following hyperparameter settings to obtain the best result on CHAIR: 

 \(\kappa_{\textsc{vis}} = 0.1\),  \(\alpha_{\textsc{neg}} = 1.0\),  \(\alpha = 1.0\), and \(\beta = 0.1\).

\textbf{LLaVA-1.5}: \(\kappa_{\textsc{tch}} = 32\), \(\alpha_{\textsc{pos}} = 0.6\).

\textbf{LLaVA-NeXT}: \(\kappa_{\textsc{tch}} = 32\),  \(\alpha_{\textsc{pos}} = 0.7\).

\textbf{Phi2-SigLIP}: \(\kappa_{\textsc{tch}} = 128\), \(\alpha_{\textsc{pos}} = 0.8\).



    

\section{Visualization Example}
\label{sec:vis_example}

\begin{figure}
  \centering
   \includegraphics[width=1.0\linewidth]{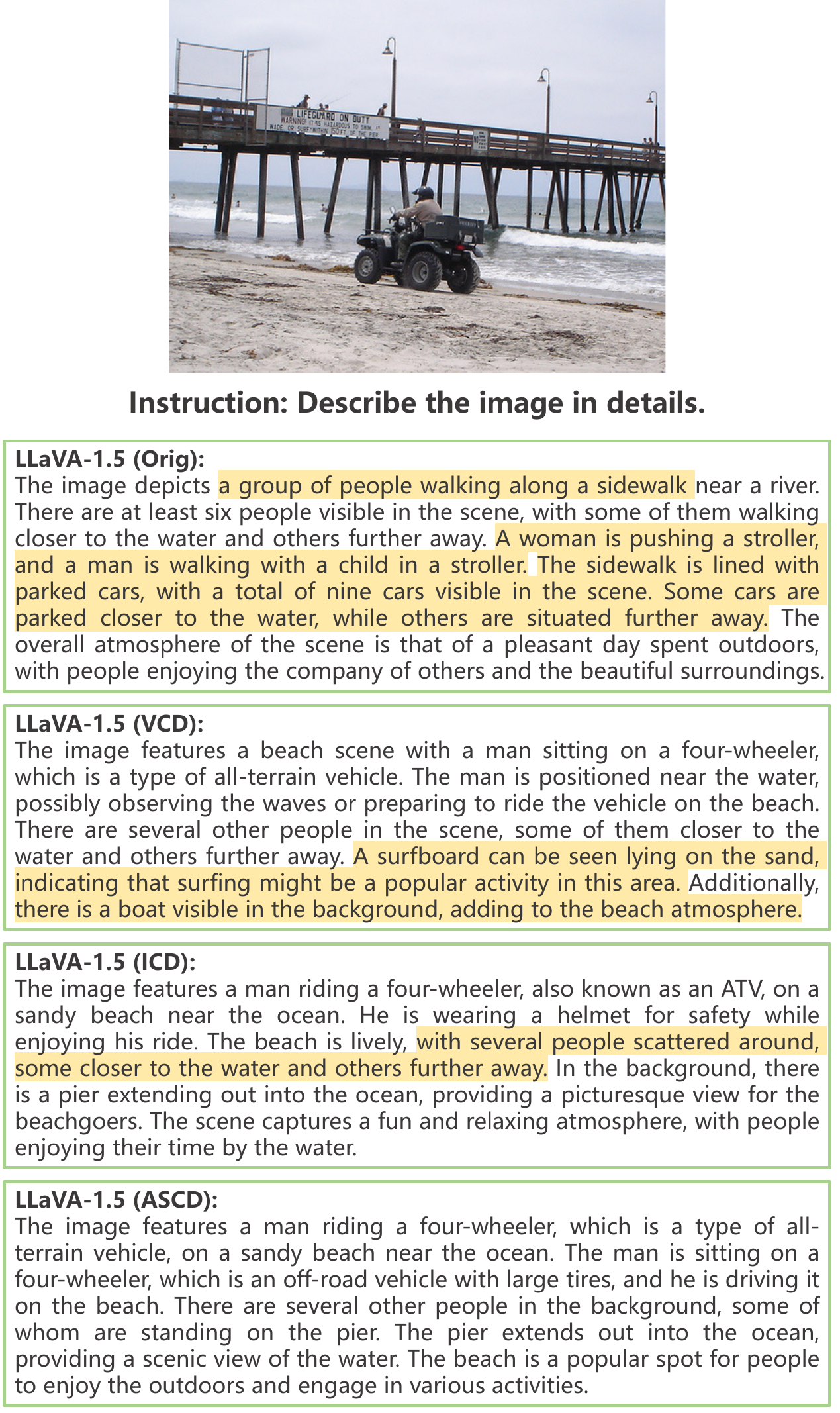}

   \caption{Original LLaVA-1.5 drifts completely, describing a riverside sidewalk, parked cars, and a stroller that do not exist in the image. VCD and ICD correctly identify the beach and ATV but still hallucinate extra objects such as a surfboard and a boat (highlighted in yellow). Our ASCD caption mentions only entities that are actually present and omits non-existent items, illustrating the improved visual grounding achieved through attention-steerable contrastive decoding.}
   \label{fig:qual_vqa}
\end{figure}

Figure \ref{fig:qual_vqa} compares the image–caption pairs produced by the original LLaVA-1.5, its VCD and ICD variants, and our \mbox{ASCD} (all with greedy decoding).  

\begin{table*}[t]
\centering

\setlength{\tabcolsep}{3pt}   

\begin{minipage}[t]{0.24\linewidth}\centering
\begin{tabular}{ccc}
\toprule
$\alpha_{\text{pos}}$ & Acc & F1 \\ \midrule
Orig & 85.4 & 84.1 \\ 0.1 & 86.0 & 85.1 \\ 0.2 & 85.9 & 85.1 \\
0.3 & 85.8 & 85.0 \\ 0.4 & 86.0 & 85.3 \\ 0.5 & 86.3 & 85.9 \\
0.6 & \textbf{86.6} & \textbf{86.3} \\ 0.7 & 85.2 & 84.2 \\
\bottomrule
\end{tabular}
\end{minipage}\hfill
\begin{minipage}[t]{0.24\linewidth}\centering
\begin{tabular}{ccc}
\toprule
$\alpha$ & Acc & F1 \\ \midrule
Orig & 85.4 & 84.1 \\ 0.25 & 86.1 & 85.3 \\ 0.50 & 86.4 & 85.9 \\
0.75 & \textbf{86.7} & \textbf{86.4} \\ 1.00 & 86.6 & 86.3 \\
1.25 & 86.5 & 86.3 \\ 1.50 & 86.6 & 86.4 \\ 1.75 & 86.6 & 86.4 \\
2.00 & 86.6 & 86.4 \\
\bottomrule
\end{tabular}
\end{minipage}\hfill
\begin{minipage}[t]{0.24\linewidth}\centering
\begin{tabular}{ccc}
\toprule
$\beta$ & Acc & F1 \\ \midrule
Orig & 85.4 & 84.1 \\ 0.2 & 86.1 & 86.0 \\ 0.4 & 86.4 & 86.2 \\
0.5 & \textbf{86.6} & \textbf{86.3} \\ 0.6 & 86.6 & 86.2 \\
0.8 & 86.3 & 85.7 \\ 1.0 & 85.7 & 84.7 \\
\bottomrule
\end{tabular}
\end{minipage}\hfill
\begin{minipage}[t]{0.24\linewidth}\centering
\begin{tabular}{ccc}
\toprule
$\kappa_{\text{TCH}}$ & Acc & F1 \\ \midrule
Orig & 85.4 & 84.1 \\ 16 & 86.3 & 85.7 \\ 32 & \textbf{86.6} & \textbf{86.3} \\
64 & 86.6 & 86.3 \\ 128 & 86.5 & 86.2 \\ 256 & 86.3 & 85.7 \\ 512 & 86.0 & 85.4 \\
\bottomrule
\end{tabular}
\end{minipage}

\caption{Ablation on key hyper-parameters (LLaVA-1.5 7B on POPE).  
Each block varies one hyper-parameter while fixing the others; bold marks the best setting per block.}
\label{tab:ablation_hyper}
\end{table*}

\section{Runtime and Memory Overhead}
\label{app:runtime}

\paragraph{Motivation.}
Attention steering operates on internal matrices at generation time and could in principle introduce non-trivial latency or memory footprints.
To verify that this is not a practical bottleneck, we benchmark ASCD against the two mainstream contrastive-decoding baselines—\textbf{VCD} and \textbf{ICD}—using the original LLaVA-1.5-7B model.

\paragraph{Setup.}
For a single A100-80G we measure (i) wall-clock latency to produce an answer for a \(224\times224
\) image–question pair and (ii) peak GPU memory, under three attention back-ends: the default PyTorch \emph{eager} kernel, fused \emph{SDPA}, and \emph{FlashAttention 2}.

\begin{table}[ht]
\centering
\small
\setlength{\tabcolsep}{5pt}
\begin{tabular}{lcccc}
\toprule
\textbf{Metric} & Orig & VCD & ICD & ASCD \\
\midrule
Latency — eager (s)       & 0.09 & 0.17 & 0.17 & 0.16 \\
Peak VRAM — eager (GB)    & 14.52 & 15.00 & 15.17 & 15.15 \\
\addlinespace
Latency — SDPA (s)        & 0.08 & 0.16 & 0.16 & — \\
Peak VRAM — SDPA (GB)     & 14.52 & 15.15 & 15.17 & — \\
\addlinespace
Latency — FlashAttn (s)   & 0.08 & 0.15 & 0.15 & — \\
Peak VRAM — FlashAttn (GB)& 14.52 & 15.00 & 15.17 & — \\
\bottomrule
\end{tabular}
\caption{\textbf{Inference cost of different decoding strategies} on LLaVA-1.5-7B.  
Dashes indicate that our current ASCD prototype can be integrated only in the eager path.}
\label{tab:runtime}
\end{table}

\paragraph{Observation.}
In \ref{tab:runtime}, ASCD adds at most \(0.01\) s latency and \(\sim\)0.6 GB VRAM over the unmodified model, while VCD/ICD incur larger overhead due to image perturbations or prompt inflation.  
Thus, the selective head-level edits of ASCD preserve virtually the same inference efficiency, confirming that attention steering is a practical choice for real-time deployment.

\section{Ablation Study of Hyperparameters}
\label{sec:ablation_study_hyperparam}

\paragraph{Hyper-parameter robustness.}
Table \ref{tab:ablation_hyper} sweeps the four controllable hyper-parameters of ASCD—positive-steer weight $\alpha_{\text{pos}}$, contrastive weight $\alpha$, truncation threshold $\beta$, and the number of text-centric heads $\kappa_{\text{TCH}}$—on the POPE benchmark with LLaVA-1.5-7B, whose mean accuracy and F1 jointly capture hallucination-reduction and VQA quality.  Across the entire search range each variant \emph{consistently surpasses} the untuned baseline, and the performance curve is flat around the optima (bold).  The result indicates that ASCD is insensitive to moderate hyper-parameter changes and can be deployed with minimal tuning effort.

\citep{bi2025cot,bi2025llavasteeringvisualinstruction,bi2025prism}

\end{document}